\documentclass[journal]{IEEEtran}
\usepackage{cite}
\usepackage{amsmath,amssymb,amsfonts}
\usepackage{algorithmic}
\usepackage{graphicx}
\usepackage{textcomp}
\usepackage{xcolor}
\usepackage{verbatim}
\usepackage{booktabs}
\usepackage{array}
\usepackage{bigints}
\usepackage[caption=false,font=footnotesize]{subfig}
\usepackage{fixltx2e}
\usepackage{stfloats}
\usepackage{url}
\usepackage{enumitem}
\usepackage{todonotes}

\DeclareMathOperator{\Var}{Var}

\begin{document}
%
\title{Seeding for Success: Skill and Stochasticity in Tabletop Games}


\author{\IEEEauthorblockN{James Goodman, Diego Perez-Liebana, Simon Lucas}
\IEEEauthorblockA{{Game AI Research Group} \\
\textit{Queen Mary University of London}\\
james.goodman, diego.perez, simon.lucas@qmul.ac.uk}

}


\IEEEoverridecommandlockouts
\IEEEpubid{\makebox[\columnwidth]{ 979-8-3503-5067-8/24/\$31.00~\copyright2024 IEEE \hfill} 
\hspace{\columnsep}\makebox[\columnwidth]{ }}

\maketitle
\IEEEpubidadjcol

\begin{abstract}
Games often incorporate random elements in the form of dice or shuffled card decks.
This randomness is a key contributor to the player experience and the variety of game situations encountered. 
There is a tension between a level of randomness that makes the game interesting and contributes to the player enjoyment of a game, and a level at which the outcome itself is effectively random and the game becomes dull. The optimal level for a game will depend on the design goals and target audience.
We introduce a new technique to quantify the level of randomness in game outcome and use it to compare 15 tabletop games and disentangle the different contributions to the overall randomness from specific parts of some games. 
We further explore the interaction between game randomness and player skill, and how this innate randomness can affect error analysis in common game experiments.

\end{abstract}


%
\IEEEpeerreviewmaketitle

\section{Introduction}\label{sect:Intro}

Games can vary significantly in their levels of randomness. Perfect information games like Chess eliminate chance entirely. Others, such as Poker and Bridge, introduce randomness through the initial deal of cards. This randomness, coupled with hidden information, contributes to the unpredictable nature of these games.
Both types of games have enjoyed enduring popularity, demonstrating that randomness is neither a prerequisite nor a detriment to a game's success.

In contemporary game design, particularly tabletop games, randomness, whether through dice, card shuffling, or other mechanisms, is a fundamental design tool~\cite{schell_art_2008,elias_characteristics_2012,woods_eurogames_2012}.
Elias et al. (2012) identify several reasons for incorporating indeterminacy, including enhancing gameplay variety (as seen in the random card deals of Bridge), broadening the appeal to diverse players (allowing less experienced players to compete in family games), and increasing the excitement of the unknown (like the reveal of a hidden card in Poker or a doubling throw in Backgammon)~\cite{elias_characteristics_2012}.
However, excessive or poorly implemented randomness can undermine a game's appeal. A game like Candyland or Snakes and Ladders, where player skill has little impact on the outcome, serve as extreme examples.

As Elias et al. point out, quantifying randomness in a game is challenging. A single dice roll can introduce randomness, but multiple rolls may lead to more predictable outcomes due to averaging effects.
Complaints that, ``the dice were against me", or ``I was just unlucky" are common at the gaming table.
In this work we seek to objectively measure the impact of the randomness of a game on outcomes, and the extent to which these complaints might have a solid foundation.

Costikyan 2013 analyses eleven different types of `uncertainty' in a game\cite{costikyan_uncertainty_2013}. 
This includes uncertainty in an opponent's response or in a player's own physical or cognitive capabilities, and is a much broader discussion of interesting forms of game uncertainty from a design perspective. 
This work deals purely with \emph{randomness} as one of these eleven types; the uncertainty in outcome of a game due to random number generation as part of the game rules, such as dice rolls or card shuffles.

We look at the effect of stochasticity in a number of different tabletop board games to quantify the impact of randomness in each game.
It builds on previous work published at the IEEE Conference on Games in 2024~\cite{goodman_measuring_2024}. This asked two key research questions:
\begin{enumerate}
\item Does the randomness integral to the game rules affect the result and if so, how big is this effect? 
\item If this randomness does have a significant effect, can the different contributions be disentangled and attributed to different sources of randomness in the game, such the shuffle of specific decks of cards, or an initial randomised board set up.
\end{enumerate}
The main results from this work are re-presented here and extended in three areas:
\begin{enumerate}[resume*]
    \item How does the skill of an agent interact with randomness in a game? The original work used agents of a single skill level. Experiments are now added in which agent skill is varied across about two orders of magnitude. This tests the extent to which greater skill mitigates or exacerbates the impact of randomness, and how this varies across games.
    Elias et al. talk about how player skill interacts with randomness to mitigate bad luck, and efficiently exploit good luck as a positive feature of game design~\cite{elias_characteristics_2012}. The new experiments concretely test this prediction.
    \item Does the randomness of a game impact error analysis? Is it possible to reduce the error bounds with knowledge of the impact of random seeds by pairing games. 
    \item How important is randomness in the game of Catan. In the original work the results for Catan indicated no impact of randomness on the game outcome despite clear evidence in the literature and game forums to the contrary. 
    It was speculated that this was due to the weak level of the agents used, and the results are revisited with better Catan-specific agents.
\end{enumerate}

\section{Background}

\subsection{TAG and Games}
The games analysed are commercial tabletop board or card games implemented on the Tabletop Games Framework (TAG) for ease of comparison~\cite{gaina_tag_2020}.
The code is publicly accessible at \url{https://github.com/GAIGResearch/TabletopGames}. 

The detail of the games included in the analysis (see Section~\ref{sect:Method}) is omitted for space and summaries of them are available at \url{www.boardgamegeek.com}. 
Four games are analysed in more detail in Section~\ref{sect:MethodDisentangle}, and outlined below. These are described in more detail in \cite{goodman_measuring_2024}:

\begin{itemize}[leftmargin=*]
    \item \textbf{Seven Wonders}. Players draft cards to build a tableau, interacting with neighbours by passing cards and buying resources. Each player has a randomly allocated Wonder with unique special abilities. There are three card decks that are shuffled once at the start of the game.
    \item \textbf{Colt Express}. Players plan train robberies by playing cards with partial visibility to the other players. The train layout, round order, and character special abilities are all randomly determined at the start of the game. Each player also shuffles their deck at the start of each round.
    \item \textbf{Dominion}. Players build their deck through card purchases to create an `engine' to acquire victory point cards in the late game. The fixed initial deck for each player is shuffled, and a player shuffles their discard deck to create a new draw deck intermittently through the game.
    \item \textbf{Catan}. Players roll dice every turn to gain resources (affecting all players) and use these to construct roads and settlements (worth points) on a randomised hex map. Different hexes provide different resources with different frequencies.
\end{itemize}

\subsection{MCTS}\label{sect:MCTS}
Monte Carlo Tree Search (MCTS) has been successful in many games~\cite{browne_survey_2012}. It has been adapted to imperfect information games with Information Set MCTS~\cite{cowling_information_2012}, which is the version used here.
MCTS builds an asymmetric game tree with partially random trajectories through the game, and then recommends the action with the best reward. The reader is referred to the cited papers for details of the algorithm.

MCTS is not core to the contributions of this paper, and any non-deterministic planning or learning algorithm could be used.
For the purposes of the discussion in Section~\ref{sect:Method} it is important to note that MCTS is a stochastic algorithm and will not always make the same action choice from a position. An MCTS agent is always initialised with a random seed.

\subsection{Player skill and game stochasticity}\label{sect:SkillBG}
Arnault and Barbie are playing Chess. Barbie is better than Arnault, but she will not win every game.
Chess is a deterministic game, so this variability in outcome must come from the players themselves. Perhaps Arnault decides to play the Queen's Gambit opening, with which Barbie is very familiar.
At a more tactical level, human players make mistakes. On Turn 14 Arnault may forget about the latent fork on his Queen 
as he enthusiastically takes the offensive, 
despite having spent time thinking about the possibility on Turn 9.
The result is that between two \emph{human} players, the result of the game is probabilistic, even if the game itself is deterministic.

Now consider two AI players, Alpha and Beta. If these algorithms are themselves deterministic, such as minimax search to different fixed depths, then we expect every game of Chess to give the same result and exactly the same moves. 

If however these players are algorithmically stochastic, as in MCTS, then we get back to a more `human'-like environment. 
This assumes different random seeds are used for the agents in each game; if the same seed is used each time, then we have a fixed outcome as with purely deterministic agents.
This enables our approach of measuring the win rate over 1000 games in which the random seed is fixed describe in more detail in Section~\ref{sect:Method}.

\section{Previous Work}

Selecting specific seeds in games to define the setup is a standard tactic and is core in Perfect Information Monte Carlo, in which N possible samples of an Imperfect Information game are each solved using minimax search or other technique, and then the results amalgamated to make a decision in the real, unknown game~\cite{ginsberg_gib:_2001}.

It can also be used to provide a sample of games as `easy' or `difficult' for AI agents \cite{sfikas_collaborative_2020}.
This also holds for human play; in Duplicate Bridge identical deals of cards are played by all participants so that their skill can be fairly compared. A pair may `lose' a hand, but gain points because in relative terms they lost less than other pairs~\cite{hosch_duplicate_2006}.

Different levels of randomness are appropriate for different audiences and target experiences. 
In many cases a game can be wildly `unfair' due to this randomness and still meet its design goals~\cite{olotka_fair_2011}. 
Some Poker deals for example may be winnable without much skill, but this averages out over many  hands~\cite{garfield_design_2011}.
In other cases this issue can become detrimental with more experienced players, who may seek more `balanced' versions. One example of this is Catan, where the initial random board layout can give a significant benefit to the first player to pick the location for their starting settlement~\cite{schreiber_settlers_2011}. This does not stop Catan being a very successful game with over 40 million copies sold and a competitive World Championship series~\cite{teuber_about_1995}.

Varying the seed used by the MCTS player has also been used to provide a number of functionally different players, despite using exactly the same algorithm~\cite{czarnecki_real_2020,st-pierre_nash_2014}. 
Analysis of this in 2-player perfect information games shows that the win rate of individual MCTS seeds (against random opponent seeds) 
can vary quite dramatically, and the distribution of this variance can be plotted~\cite{st-pierre_nash_2014,liu_fast_2016}. This distributional consideration is perhaps closest to this work, but we fix the \emph{Game} seeds instead of the \emph{MCTS} seeds.

\section{Error Analysis}
Stochastic agents in stochastic games provide two sources of noise. 
This complicates the standard distributional assumptions for confidence bounds or p-values on win rates in games.
The standard model is that each game outcome samples a Bernouilli($p$) distribution. Player A wins with probability $p$ and loses with probability $1-p$.
If $N$ games are run between two agents and $p$ is constant for all games then the number of Wins of Player A is sampled from a Binomial($N, p$) distribution. This allows exact confidence intervals, $B$, on the estimate $\hat{p} = (\text{Wins by A}) / N \pm B$. 

An underlying assumption is that $p$ is fixed for all games. As per the earlier discussion, this is often not true and will vary with the random seed.
For deterministic agents each individual game either has $p=0$, or $p=1$; re-running with the same game random seed will give a deterministic outcome.
Let $p_\mu$ be the average win rate across all possible seeds. 
Sampling a new random seed for each of $N$ games
on \emph{average} may give, say, $p_\mu=0.6$, and the number of wins in $N$ games
is still a Binomial($N, p_\mu$) distribution.

This can seem counter-intuitive given there are two sources of stochasticity: first a random generation of the true underlying probability of a win, then a second random roll to see if the win actually occurs.
However, this makes no difference and  the variance of the outcome of one game is the same as the variance of a Bernoulli draw ($p(1-p)$).

Each individual game has one of a discrete number of outcomes (Win, Lose; or Win, Lose, Draw) with an assigned value. The two (or more) sources of stochasticity that affect this ultimate probability do not affect the calculation; it is only this final probability that determines the variance of the result (and hence any confidence intervals). Each game is \emph{still} a single Bernoulli draw for the net $p$ as proved below.

Let $X_i \in \{0, 1\}$ be the outcome of the $i$th game.
Let $Pr(X_i = 1) = p_i \sim f(\xi) : [0, 1] \to [0, 1]$, i.e. the probability of winning each game is individually sampled from an arbitrarily complex probability distribution $f(\xi)$, such as the sequence of rolls imagined above. 
Let $f(\xi)$, and hence $p_i$, have mean $p_\mu$. 
To obtain the expectation of $X$ we integrate over $\xi$, the unknown (and arbitrary) latent parameter defining the distribution of $p_i$. From the definition of $\mathbb{E}[X]$:
\begin{align}
    \mathbb{E}[X_i] &= \int\limits \text{Pr}(X_i = 1 | \xi) \, f(\xi) \, d\xi \\
    &= \int\limits_0^1 p_i \, f(\xi) \, d\xi \\
    &= \mathbb{E}[p_i] = p_\mu
\end{align}
I.e. the mean is unchanged regardless of the form of $f(\xi)$. Similarly for the variance:
\begin{align}
    \Var(X_i) &= \int\limits_0^1 f(\xi) (X_i - p_\mu)^2 d\xi \\
    &= \int\limits_0^1 \left(X_i^2 - 2X_ip_\mu + p_\mu^2\right) f(\xi) d\xi \label{eqn5}\\
    &= \mathbb{E}\left[\left(X_i^2 - 2X_ip_\mu + p_\mu^2\right)\right]   \label{eqn6}\\
    &= \mathbb{E}[p_i - 2p_i p_\mu + p_\mu^2] \label{eqn7} \\
    &= p_\mu(1 - p_\mu)
\end{align}
This last line is the variance of a Bernoulli distribution with probability $p_\mu$.
A key simplification is that $X$ is a binary outcome. Hence from (\ref{eqn6}) to (\ref{eqn7}) we can use $\mathbb{E}[X_i^2] = \mathbb{E}[Pr(X_i = 1)] = p_i$.
If $X$ were a more complicated outcome, such as a score, then this result would not hold and the variance would be notably higher due to the extra sources of stochasticity, even though the mean is unchanged. 
Drawn games being worth 0.5 contravene requirement for a simple binary outcome so this result is appropriate for games where draws are infrequent.

With the important proviso that the binary win rate is the outcome of interest the outcome of $N$ independent games is hence a sample from a Binomial($N$, $p_\mu$) distribution regardless of the details of the distribution of $p_i$. Using the standard error bounds from a Binomial distribution, or its Normal approximation for large $N$ remains valid.

\subsection{Re-using random seeds to reduce variance}

Despite this, the ability to control the stochasticity in the game  by re-using game seeds can be used to reduce the variance of the experimental outcome.
In a 2-player game this is done by playing the game twice, starting with the same seed each time but with the players switching positions. 
In the case of Poker, this means each player gets to play the same cards in hand and on the table. 

The two sources of stochasticity apply to each game \emph{individually}, which remains a Bernoulli($p$) draw, assuming we have 0 for a loss and 1 for a win. However the two mirrored games are now correlated, and the net outcome is \textbf{not} a Binomial($2, p$) distribution as the iid assumption is breached.

If the two agents are identical the probability that the first player wins is $p_i^1$, and the second player wins with probability $1 - p_i^1$ (disregarding draws for simplification).
Now consider the expectation and variance of the pair of games, where $X_2 \in \{0, 1, 2\}$ is the number of games won by a given player (who plays first in the first game and second in the second):

\begin{align}
    \mathbb{E}[X_{2, i}] &= \int\limits_0^1 p_i^1 + (1 - p_i^1) f(\xi) d\xi \\
    &=  \int\limits_0^1 f(\xi) d\xi \quad= 1 \label{eqn11}
\end{align}
Regardless of random seed, this will balance out if two mirror games are played and each player is expected to win one. 
In the more general case where draws are possible, $\mathbb{E}[X_{2, i}] = \mathbb{E}[p_i^1 + p_i^2] = 2 p_\mu$. 

$X_2$ is the sum of two Bernoulli distributions, one with probability $p_i^1$, and one with probability $p_i^2 = (1 - p_i^1)$.
Calculating the variance from first principles gives:
\begin{align}
    \Var(X_{2, i}) &=   \int\limits_0^1 (X_{2, i} - 1)^2 \;  f(\xi) \; d\xi \\
    &=  \int\limits_0^1 \left(X_{2, i}^2 - 2 X_{2, i} + 1 \right) \; f(\xi) \, d\xi \\
    &=  \mathbb{E}[X_{2, i}^2] - 2 \mathbb{E}_\xi[X_{2, i}] + 1  
\end{align}
\begin{align}
  \Var(X_{2, i}) &=  \mathbb{E}[X_{2, i}^2] - 1 \label{eqn15} \\
    &=  \mathbb{E}[4p_i^1(1 - p_i^1) + (p_i^1)^2 + (1 - p_i^1)^2 ] - 1 \label{eqn16} \\
    &=  \mathbb{E}[2p_i^1(1 - p_i^1)] \label{eqn17}
\end{align}
(\ref{eqn15}) follows using (\ref{eqn11}), and (\ref{eqn16}) enumerates the possible values of $X_{2,i}$ (2, 1 or 0) multiplied by their respective probabilities.
This result is, as expected, the sum of two Bernoulli distributions: Ber($p_i^1$) + Ber($1-p_i^1$).

\begin{figure}[h]
    \centering
    \includegraphics[scale=0.5]{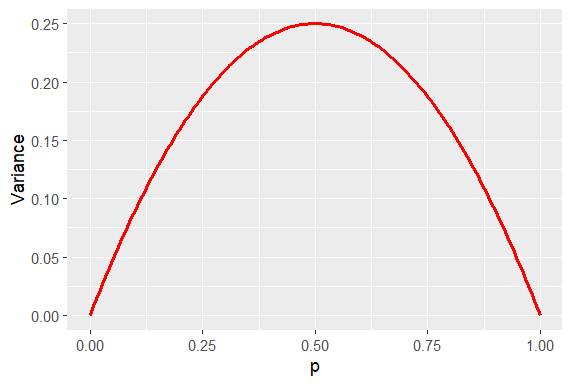}
    \caption{Plot of variance ($p(1-p)$) shows the variance peaks at $p=0.5$, and is zero for $p \in \{0, 1\}$. Any change in the distribution of $p$ from a constant $p=0.5$ will therefore reduce the variance from this peak.}
    \label{fig:varSimple}
\end{figure}

The non-linearity of (\ref{eqn17}) compared to (\ref{eqn7}) with the term in $(p_i^1)^2$ means this has no closed form, and is dependent on the precise details of the distribution $f(\xi)$.
However, it does have a maximum at $p_i^1 = 0.5$, and knowledge of the distribution of $p_i^1$ will reduce the variance in $X_2$ (Figure~\ref{fig:varSimple}).

Only in the case that every game is perfectly fair with $p_i = 0.5$ will running mirror games not reduce the variance. 
These results assume that $f(\xi)$, the distribution of the results for different seeds, is known, which is generally not the case. 
One post hoc approach is to take the empirical distributions of $p_i^1$ from the empirical distributions reported in Section~\ref{sect:resultsBase} and shown in Figure~\ref{fig:Sample}, and use these to calculate the relative reduction in the variance.

\begin{table}[b]
    \centering
    \begin{tabular}{lcc|cc}
     & \multicolumn{2}{c|}{Single} & \multicolumn{2}{c}{Mirrored} \\
    Game & Variance & 95\% interval & Variance & 95\% interval \\
    \toprule
     Dots + Boxes & 0.247 & 6.2\% & 0.247 & 6.2\% \\
     Dominion    & 0.249 & 6.2\%& 0.243 & 6.1\%\\
     Can't Stop & 0.248 & 6.1\% & 0.237 & 6.0\% \\
     7 Wonders & 0.231 & 5.9\% & 0.221 & 5.8\% \\
     Virus & 0.249 & 6.2\% & 0.227 & 5.9\% \\
     Love Letter & 0.250 & 6.2\% & 0.218 & 5.8\% \\
     Colt Express & 0.248 & 6.2\% & 0.220 & 5.8\% \\
     Sushi Go & 0.249 & 6.2\% & 0.204 & 5.6\% \\
     Hearts & 0.222 & 5.8\% & 0.196 & 5.4\% \\
     Stratego & 0.246 & 6.1\% & 0.222 & 5.8\% \\
     Poker & 0.249 & 6.2\% & 0.113 & 3.5\% \\
    \bottomrule
    \end{tabular}
    \caption{Reductions in Variance for running 1000 games either as mirrored-pairs across 500 random seeds, or as single games with 1000 random seeds. Variance is `per game' with a theoretical maximum of 0.250. The 95\% interval is for the estimated win rate using an exact binomial test.}
    \label{tab:rnd:error}
\end{table}

Table~\ref{tab:rnd:error} shows the reductions in variance and error bounds for the games considered here using this post hoc approach.
This sampled empirical distribution will approach the true underlying distribution asymptotically by the Law of Large Numbers, so these reductions are not precise and are upper bounds on the actual reduction.
This reduction is usually quite small, constricting the 95\% confidence interval on the mean win rate by the order of 0.1 to 0.4\% in Table~\ref{tab:rnd:error}, with no pattern visible on type of game.
Only for Poker, with its extreme bimodal distribution over seeds does this significantly affect the bounds, almost halving them in size. 

This is not efficient as it takes in this case 100,000 games to estimate the distribution $f(\xi)$, and then estimate the variance reduction on a set of 1000 games. It would clearly be better to just use the total game budget for the direct estimate.

Even if the distribution of seeds is not known, using a fixed set of seeds for \emph{comparison} of agents reduces the variance of this comparison in line with Table~\ref{tab:rnd:error} if each agent plays from each position in the game, balancing out the often significant range of first player advantages.
Mirroring games by random seeds reduces the variance in the experimental results, even if this cannot be quantified directly.

\section{Methodology}\label{sect:Method}

For each game 200 different random seeds are sampled. For each random seed 1000 games are run and the win rate of the first player measured (a draw is counted as 0.5 of a win). For each of these 1000 games the MCTS players have different random seeds. 
This gives a sample of 200 different win rates.

If the stochastic elements of the game have no net impact on the outcome then this distribution of 200 win rates (one per seed) will be tightly clustered around $p_\mu$. If the game has no first player advantage then $p_\mu = 0.5$, but this is not required.
Each of the mean win rates is the average of 1000 independent games, so  99\% confidence bounds can be calculated from a Binomial(1000, $p_\mu$) distribution. On average 1\% of the mean win rates are expected to fall outside these bounds.

Four metrics are assessed for each game:
\begin{enumerate}
    \item Entropy of the distribution of win rates. The win rates are discretised into 2\% buckets, and the Shannon entropy, $S$, of this discrete distribution calculated, $S = \sum_i - p_i \log p_i \, i \in 1..50$.
    \item The number of samples that fall outside the 99\% binomial confidence interval. If this is much larger than 0.01 then there is evidence that the outcome of the game is affected by the specific random seed.
    \item Span. The maximum win rate of the 200 samples minus the minimum win rate. This will vary from 0 (all game seeds give the same win rate), to 1.0 (seeds vary from a 100\% win rate for the first player to a 0\% win rate).
    \item Trimmed Span. To reduce sensitivity to outliers, take the central 95\% of Span, discarding the most extreme 5\%.
\end{enumerate}

All games are run for the minimum number of players they support. This is 3-players (Hearts, Seven Wonders, Catan, Puerto Rico), or 2-players for the other eleven games.
For all experiments MCTS agents with a 50ms computational budget per decision are used. The MCTS parameters are tuned separately for each game to ensure that the agent plays reasonably well.  

A game-specific heuristic function is used for Catan only, learned from data generated through expert iteration~\cite{anthony_expert_2021}. This was required to get a sufficiently high standard of skill. 
Results are reported in Sections~\ref{sect:resultsBase} and \ref{sect:ResultsCatan}.

There are a number of distinct random seeds used in any given game. These are:
\begin{enumerate}
    \item A seed used to control all game events (deck shuffles, dice throws etc). This is the seed fixed for each set of 1000 games and is akin to the chance player in OpenSpiel~\cite{lanctot_openspiel_2019}.
    \item A seed used to redeterminise the state to hide hidden information before being passed to the MCTS agent for a decision. This needs to be kept distinct from the game chance player to ensure the same sequence of future shuffles regardless of player decisions.
    \item A seed for each MCTS agent.
    The random seed of the game is not known or accessible to the agent, which makes decisions  only on current public information. 
\end{enumerate}

It is important that there is a clear separation between the random seeds that drive game actions and the random seeds that drive agent decisions, and that these random seeds drive all the variability of outcome.
This was confirmed by running several games with fixed random seeds for all games and agents and confirming that each game played out identically. (Although this is not fully guaranteed. Using a fixed time budget instead of a fixed iteration budget may mean different runs have different numbers of MCTS iterations due to operating system and garbage collection variations. This was not found to be an issue in practice.) 

\subsection{Interaction of skill and randomness}
What happens as the players in a game become more or less skilled? 
Consider the cases of `optimal' and `random' players. 
If a random seed gives one player a benefit, be it a good dice roll or a favourable hand of cards then a random agent is likely to throw away this benefit with a poor move whereas an optimal player would not. 
Real players are somewhere between these two cases and it is expected that as player skill increases the game becomes increasingly deterministic (\emph{given} the random seed) as fewer and fewer sub-optimal decisions are made that throw away the benefit.

This is similar to noting that Connect Four
is proven to be a win for the first player~\cite{allis_knowledge-based_1988}.
There is a single `seeding' of the set up (an empty board), all the information is visible, and optimal play (were an agent capable of it) would deterministically win. 
This comparison is only fully equivalent in games where results of the seed are fully visible at the start to both players. In games where this is not true, for example where future dice rolls are unknowable, even if pre-determined, there will likely be more variation in outcome.

The extent of this predicted asymptotic tendency towards determinism based on the seed will vary by game as there are large differences in how any `benefit' from a random seed manifests. For example,
\begin{enumerate}
    \item If the game has a sequence of future dice rolls, such as Can't Stop, then these future  events are unknowable and no player can condition their decisions on them. Players can only condition decisions on observed events and future expected distributions. 
    \item If the randomness is in the form of an initial shuffle and deal of a deck of cards; for example in Hearts or Poker, then the result is `baked in' at the start. In the case of Hearts there is no new randomness once a hand begins. A player knows how strong their hand is and the better hand will win in most cases if all play optimally (with exceptions for mixed strategies in finesse-like situations).
    \item Colt Express is an example that blends these two. The initial random character allocation is fully visible and players can condition all their decision on this. The shuffle of a player's hand at the end of each round is only visible once that round is reached.
    In Poker the deal of the hole cards revealed in later rounds is unknowable; an optimal player can condition their strategy on the expected distribution of these, but not the actual cards that will be drawn from the shuffled deck. 
\end{enumerate}

The experimental method of the previous section is extended to run 1000 games for each of 200 random seeds using agents of increasing skill. These skill levels are represented by the budget available to MCTS, and five values are used: 2ms, 10ms, 50ms and 250ms of thinking time, plus a purely random agent (effectively zero milliseconds of budget).
The same 200 seeds are used for each budget, and all agents in the game use the same budget.
The results are reported in Section~\ref{sect:resultsSkill}.

\subsection{Disentangling randomness}\label{sect:MethodDisentangle}
For some games, such as Poker or Hearts, there is no source of randomness apart from the shuffle of a single deck of cards at the start of the game or round.
In others there is a natural separation of `sources' of randomness. In Seven Wonders every player is dealt a Wonder board at the start of the game from a shuffled deck. This provides them with unique options, and their game is also affected by the Wonders of the other players;  players can buy resources from their immediate neighbours and some Wonders are more geared to specific strategies. 
The other source of randomness comes from the three decks of Age cards, which are each shuffled once at the start of the game. 

Additional random seeds were added to selected game implementations to measure the impact of these different sources. For example, in Seven Wonders a seed was added to control the shuffle of the Wonder boards, and a separate one to control the shuffle of the Age cards.
Two sets of experiments are then run, repeating the previous methodology, i.e. 
\begin{itemize}
    \item Sample seeds for the Wonder board shuffle (or the Age deck shuffle)
    \item For each of these seeds run 1000 games with all \emph{other} sources of randomness (game or player) initialised differently for each game.
\end{itemize}

The games selected for this approach are:
\begin{itemize}[]
    \item Seven Wonders. A separate seed for the Wonder board shuffle is introduced, and one for all the Age decks (a single seed is used by all three decks, not one each).
    \item Colt Express. Three new seeds are introduced to control each of:
    \begin{itemize}
        \item the shuffle and deal of each player's character;
        \item  the shuffle of the train carriages that make up the board; 
        \item the shuffle of the round cards.
    \end{itemize}
    \item Dominion. One new seed is introduced that controls the initial shuffle of the starting decks. This determines the first two hands of each player, which will either split the 7 Copper cards 3/4 (5 times in 6) or 2/5 (1 in 6). 
    This tests how important this initial split is.
    \item Catan. Two seeds are introduced. One to control the initial board set up and one to control dice rolls later in the game.
    \item Theme Park. This is a game in commercial development by Bright Eye Games\footnote{\url{https://www.brighteyegames.com/}}.
The game is set in a theme park, with children being taken on a day out and all wanting to go on different rides. The children will have damaging tantrums if they do not get what they want. The winner is the player who best balances their competing demands by closing time.
The game has two key decks of cards:
    \begin{itemize}
        \item A set of Person cards, of which each player receives four at the start of the game. These are public and define which rides each child wants to visit.
        \item A set of `Magic' cards that are one-off special abilities. These are face-down until drawn.
    \end{itemize}
There are no other sources of randomness or hidden information in the game One question posed during the development process is whether one or other of these two decks is too `random' and unbalancing.
\end{itemize}

\section{Results}
\begin{figure}
\centering
    \includegraphics[scale=0.3]{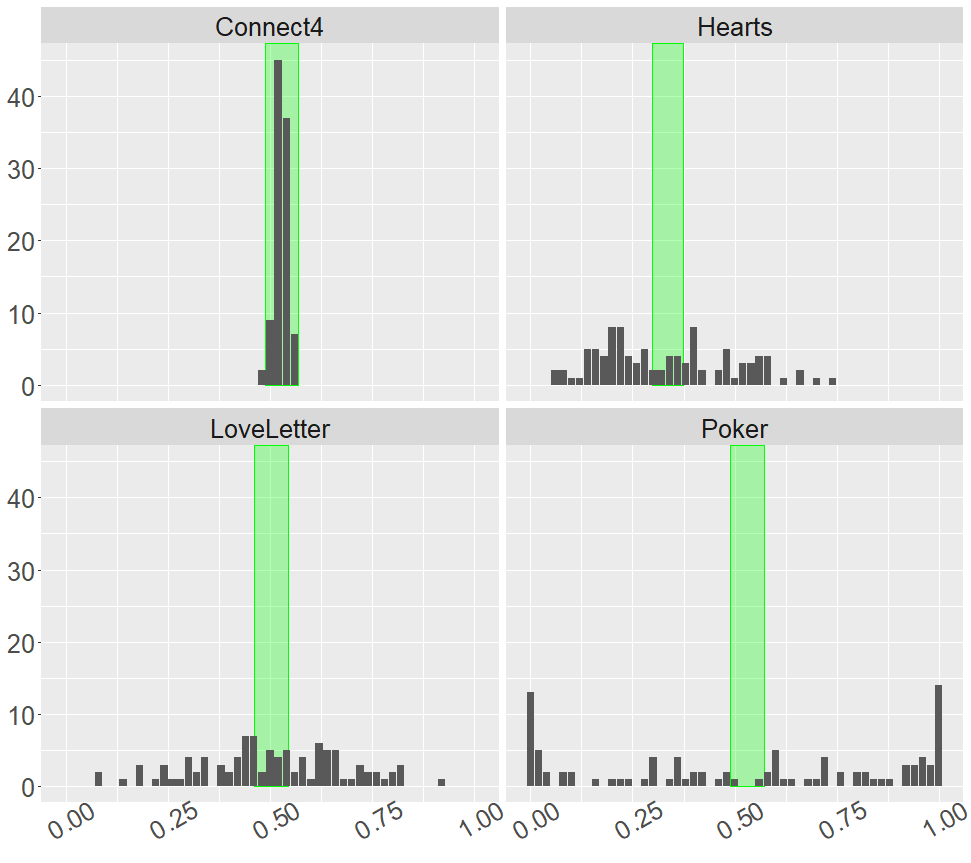}
    \caption{Random Seed plot for all 4 example games. The green shading is the 99\% confidence interval for the win rate assuming that the random seed has no effect. The x-axis is first player win rate, and the win rates for the 100 seeds are plotted as a histogram with buckets of width 2\%.}
    \label{fig:Sample}
\end{figure}

\subsection{Baseline}\label{sect:resultsBase}

Figure~\ref{fig:Sample} plots the histograms of the first player win rates for each seed for 4 of the games. If these all fall within the green-bounded 99\% confidence interval for a game then the random seed has no impact on the outcome. 
This is the case for purely deterministic games included as controls such as Connect 4.
The other three games show varying levels of impact, with Poker at the other extreme.

\begin{table}[]
    \centering
    \begin{tabular}{lccccc}
        Game & Players & Span & T-Span & Entropy & Outliers \\
        \toprule
        Dots and Boxes & 2 & 0.06 & 0.06 & 1.13 & 0.00 \\
        Diamant & 2 & 0.08 & 0.05 & 1.16 & 0.01 \\
        Connect 4& 2 & 0.08 & 0.06 & 1.21 & 0.02 \\
        Puerto Rico & 3 &  0.10 & 0.08 & 1.32 & 0.05 \\
         Dominion & 2 &  0.32 & 0.27 & 2.58 & 0.56 \\
         Can't Stop & 2 &  0.52 & 0.37 & 2.90 & 0.60 \\
         Seven Wonders & 3 &  0.48 & 0.34 & 2.89 & 0.72 \\
         Virus & 2 &  0.92 & 0.70 & 3.23 & 0.73 \\
         Love Letter & 2 &  0.99 & 0.87 & 3.69 & 0.88 \\
         Colt Express & 2 &  0.66 & 0.50 & 3.23 & 0.86 \\
         Sushi Go & 2 &  0.80 & 0.67 & 3.41 & 0.76 \\
        Hearts & 3 &  0.65 & 0.52 & 3.25 & 0.81 \\
        Stratego & 2 & 0.66 & 0.63 & 3.02 & 0.94 \\
        Poker & 2 & 1.0 & 0.99 & 3.63 & 0.95 \\
        \midrule
    \end{tabular}
    \caption{Span, Trimmed Span, Entropy and Outlier metrics for each game. Span is the difference between the best and worst seeds; Trimmed Span removes the 5\% most extreme seeds. Entropy is that of the histograms in Figure~\ref{fig:Sample}. Outliers is the proportion of seeds outside the central 99\% confidence interval.}
    \label{tab:Overall}
\end{table}

Table~\ref{tab:Overall} reports the metrics for the experiments on all 15 games.
Span, Entropy and Outliers all concur on the general pattern of which games have an outcome highly dependent on the specific random seed. 
The Span measure can be swayed by one or two outliers, as can be seen for Hearts and Love Letter in Figure~\ref{fig:Sample}. The Trimmed Span measure removes the most extreme 5\% of the results to improve robustness.

The entropy of the distribution can be misleading as the comparison of Poker and Love Letter makes clear.
In Figure~\ref{fig:Sample} Poker has the widest Span of any game, at the maximum 1.0, but it has a lower entropy than Love Letter. This is because there are two peaks in the distribution at 0\% and 100\%, which reduce the entropy of the distribution.
With a modicum of skill so that good cards are taken advantage of it is simply impossible to win (or lose) for some shuffles of the deck.
The implementation of Poker in TAG gives each player 50 chips, with a big blind of 10. Hence each game only lasts for a short number of hands (often just one if player's go All-In). This explains the high impact of the random seed.

A more detailed discussion of the results for other games, and extensions of Figure~\ref{fig:Sample} to all 15 games is in~\cite{goodman_measuring_2024}.
The most useful measures are the Trimmed Span and proportion of outliers as these are more robust than the Entropy or untrimmed Span.

\subsection{Skill and randomness}\label{sect:resultsSkill}

\begin{figure}
    \centering
    \includegraphics[width=\linewidth]{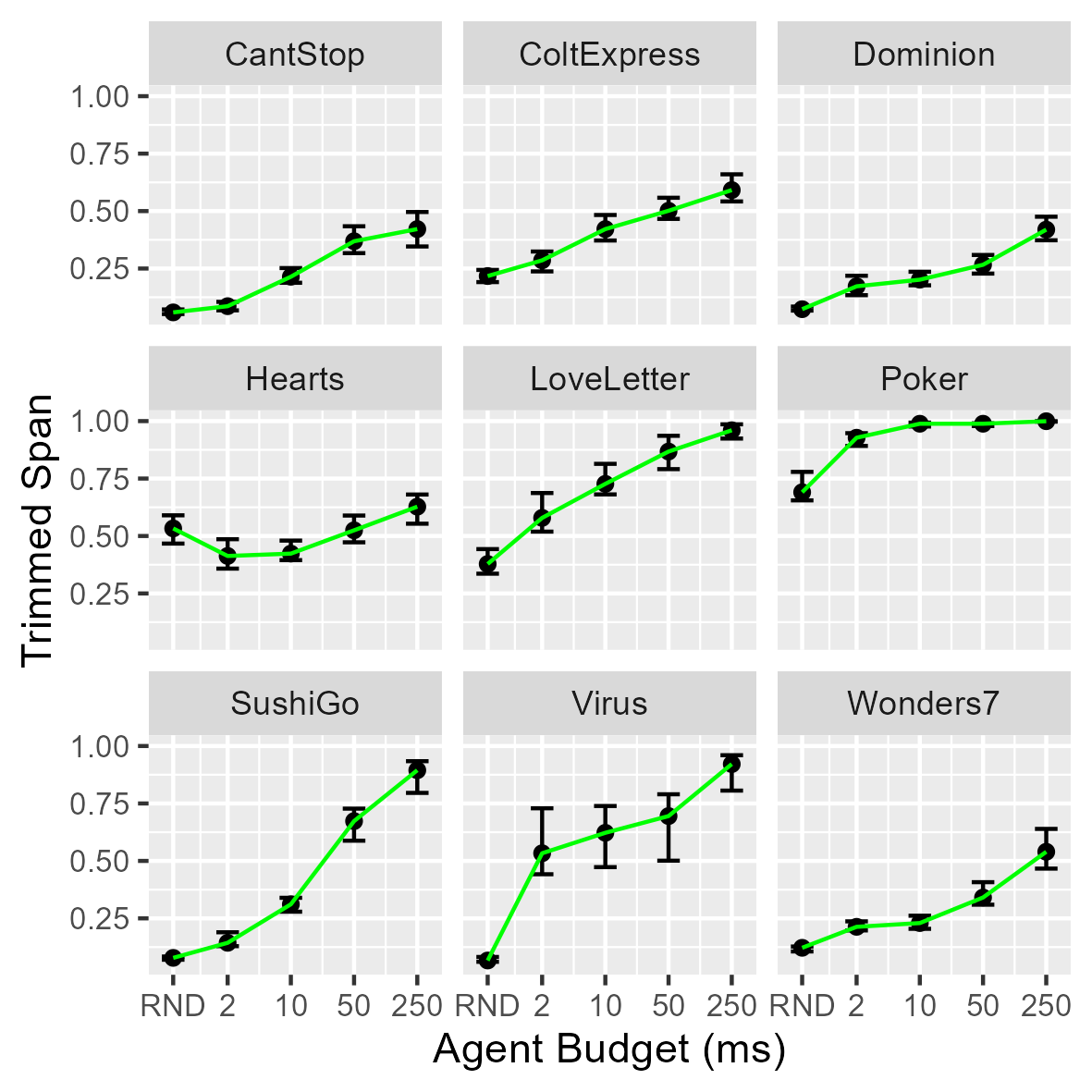}
    \caption{Impact of randomness with skill of players. Trimmed Span is reported for each of the five agent budgets. All games are played between agents using the same budget. There is a general trend for higher budget (more skilled) agents to better exploit any benefit from different random seeds.}
    \label{fig:budgetVariance}
\end{figure}

Figure~\ref{fig:budgetVariance} shows the results of the experiments between homogenous agents of increasing skill.
The error bars are 95\% confidence bounds calculated by a bootstrap on the underlying sample of 200 observations for each data point.

There is a clear pattern in all or most games for the effect of the random seed to increase as the skill of the players increases.
This is in line with the expectation that as skill increases players will more effectively exploit  benefits gained from random factors in the game.
There is also a wide variation between games, for example:
\begin{itemize}
\item Poker. This has a low branching factor as a player mostly folds, checks or raises. This is related to the highest impact of seeds on play by Random agents, as they are less likely to throw away the benefit of good cards.
\item Virus. This is arguably the lowest-skill game used but with a higher branching factor as any card in hand can be played~\cite{goodman_measuring_2024}. These combine to give a major jump in seed impact at low levels of skill (2ms of budget) that then plateaus.
\item Hearts. This is the game with the least visible trend for the impact of randomness to increase with skill. This is unexpected as skill is needed to play each hand to score well. The reason for this is not currently clear. Random seeds do have an impact, so this pattern is not because the agents are very bad at the game, in which case the impact of seeds would be consistently low.
\end{itemize}

None of these game-specific differences detract from the main pattern that the more skilled the agents, the more innate game randomness skews the expected result in any one game. 

\begin{figure}
    \centering
    \includegraphics[width=\linewidth]{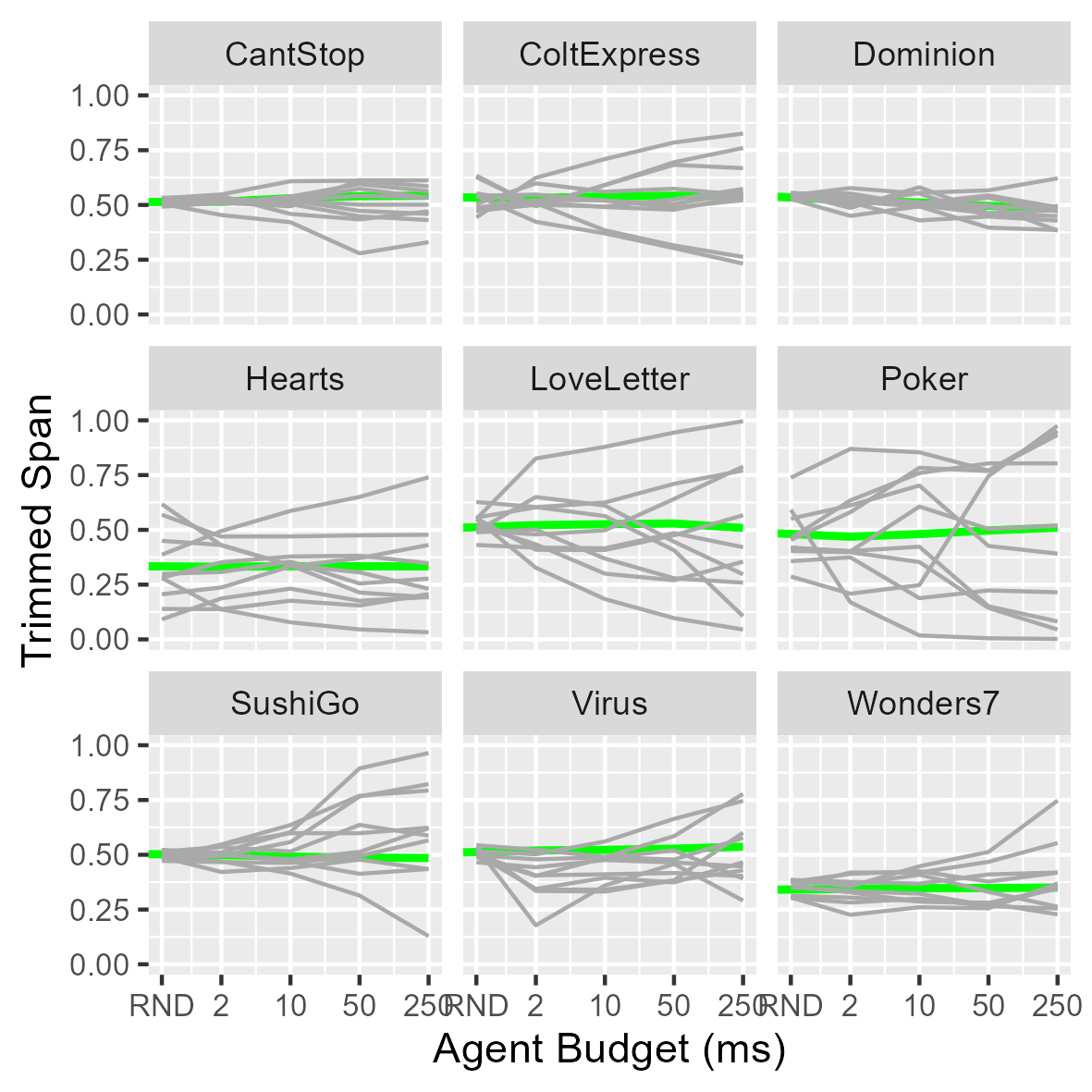}
    \caption{Individual seed results. The central green line is the average win rate of the first player (P1) as agent budget changes. The grey lines show the same lines for 10 randomly selected seeds.}
    \label{fig:seedTraces}
\end{figure}

Figure~\ref{fig:seedTraces} shows that the overall win rate (the green line) of the first player does not change at different budgets. However, the smooth transition of Figure~\ref{fig:budgetVariance} can be irregular at the individual seed. The first player win rate for most seeds increases or decreases monotonically as the agent skill increases, but there are examples where the first player has an advantage at one budget setting (say, 50ms), and much lower performance at higher and lower settings (10ms, 250ms). 
We term these `non-monotonic seeds'.

This is not expected under the assumption that increasing skill monotonically improves the ability to mitigate/exploit random game events. 
It suggests that for some game set ups there are situations where an increase in skill level 
can lead to a detection of a medium-term bonus, which at higher skill levels is discounted, or outweighed by a longer-term benefit from some other action. 
It is also possible that first player (dis)advantage may be non-monotonically variable with player skill in the same way, and could partially explain the results.

Because these are games of imperfect information, the medium-term bonus may in fact be real \emph{for this seed} even if it is not when averaged over all possible game states compatible with the current information set (recall that the MCTS agent does not know the game seed being used, and cannot use it in planning).

One example in Poker could be a medium strength hand. At 50ms of analysis an agent may conclude incorrectly that betting strongly on the hand is a good idea based on short-term planning with a finite stochastic sample of possible opponent cards. This happens to be correct if the opponent has a weak hand.
When result are averaged across all the possible states the game could be in the 250ms agent is better, but for any specific setting of the unknown cards in the opponent hand and draw deck the 50ms agent's poorer strategy might win more games.

This is related to a k-level trap in minimax search\cite{ramanujan_adversarial_2010}. A k-level trap is where search to a depth of $k$ (with some heuristic value function) chooses a best move that is in fact a guaranteed loss (or poor move more generally) if the search were taken to depth $k+1$ or greater.
A subtle difference to a k-level trap is that this trap is seed-specific, and hence may be unknowable even with full search of the game tree if the seed determines the outcome of future events.
The frequency of this reversal may be indicative of an interesting aspect of the game, depending on the design goals.

Table~\ref{tab:seedTraces} summarises the proportion of the seeds for each game with this non-monotonic variation. This counts all seeds for which the first player the win rate at 10ms or 50ms is significantly (with 99\% confidence) above or below \emph{both} the bracketing budgets.
Given the 99\% confidence interval used for each of 2 tests, about 2\% of all seeds would be highlighted if there were no pattern in the win rates with budget.
Seven Wonders and Colt Express therefore do not exhibit this reversal trait.
Deeper analysis of the individual non-monotonic seeds would be useful to investigate interesting game set ups and to probe deficiencies in the algorithm used by the agent.
\begin{table}[]
    \centering
    \begin{tabular}{lccc}
    \toprule
    Game & Non-monotonic & At 10ms & At 50ms \\
    \midrule
    Dominion & 0.17 & 0.11 & 0.12 \\
    Poker & 0.16 & 0.11 & 0.05 \\
    Virus & 0.16 & 0.07 & 0.12 \\
    Love Letter & 0.13 & 0.06 & 0.10 \\
    Sushi Go & 0.09 & 0.03 & 0.06 \\
    Hearts & 0.06 & 0.04 & 0.02 \\
    Can't Stop & 0.06 & 0.02 & 0.04 \\
    Colt Express & 0.03 & 0.00 & 0.03 \\
    Seven Wonders & 0.02 & 0.01 & 0.00 \\
    \midrule
    \end{tabular}
    \caption{Percentage of non-monotonic seeds as MCTS budget increases. Two tests are made, one at 10ms and one at 50ms. The first column shows the total proportion of non-monotonic seeds as some are in both categories. }
    \label{tab:seedTraces}
\end{table}

The aggregate data in Figure~\ref{fig:budgetVariance} is compatible with the claim from Elias et al.~\cite{elias_characteristics_2012} that good game design allows skill to mitigate obstacles or exploit opportunities thrown up by chance events.
The individual traces of Figure~\ref{fig:seedTraces} shows that this can be highly variable to specific events in some games.

\subsection{Disentangling sources of randomness}
\begin{figure}[t]
    \centering
    \includegraphics[scale=0.8]{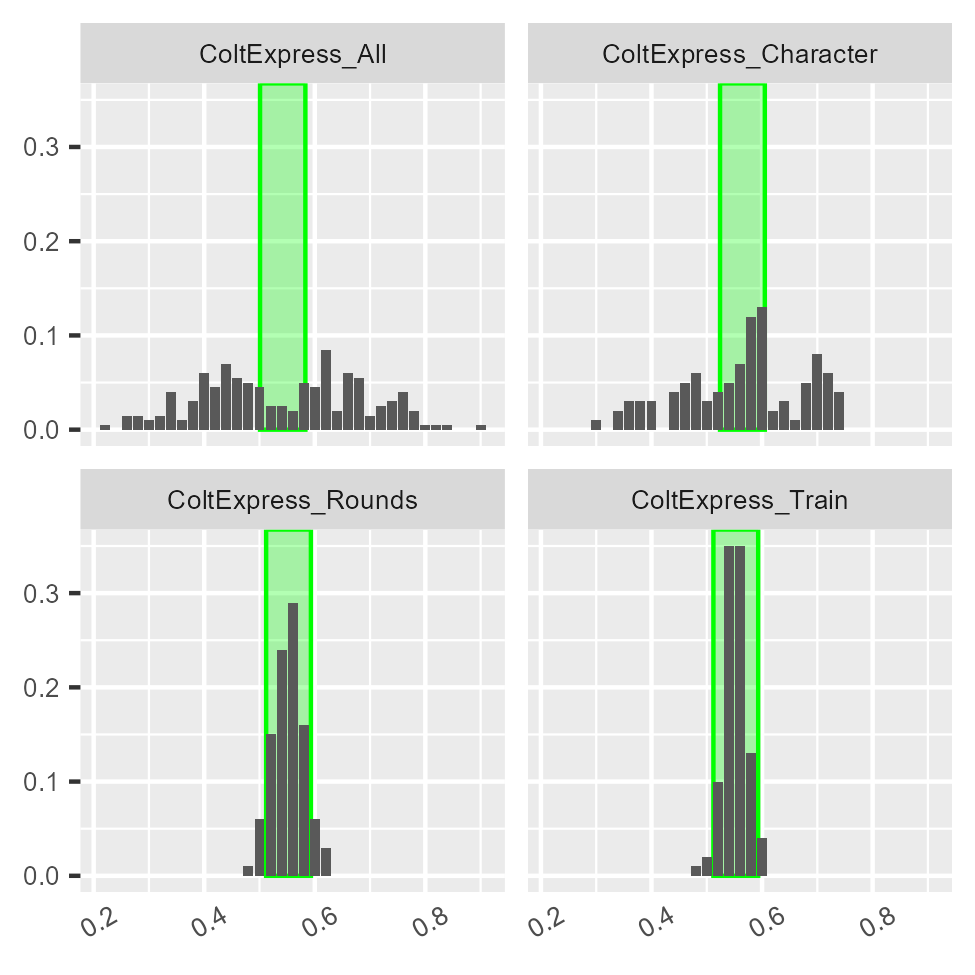}
    \caption{Comparison of the impact of initial Character, Round and Train shuffles for Colt Express. Of these three factors, only the Character randomisation has a major impact on game outcome.}
    \label{fig:Disentangling}
\end{figure}

Table~\ref{tab:Disentangling} summarises result of disentangling the contribution of different sources of randomness. Figure~\ref{fig:Disentangling} shows the details for Colt Express; see \cite{goodman_measuring_2024} for detail on the other games.

\begin{table}[]
    \centering
    \begin{tabular}{lccccc}
        Game & Constant &Span & T-Span & Entropy & Outliers \\
        \toprule
        Seven Wonders & All & 0.48 & 0.34 & 2.89 & 0.72 \\
        Seven Wonders & Wonder Board & 0.24 & 0.21 & 2.43 & 0.55 \\
        Seven Wonders & Age Cards & 0.26 & 0.18 & 2.23 & 0.39 \\
        Dominion & - & 0.32 & 0.27 & 2.58 & 0.56 \\
        Dominion & Initial Shuffle & 0.24 & 0.18 & 2.30 & 0.43 \\
        Colt Express & All & 0.66 & 0.50 & 3.30 & 0.86 \\
        Colt Express & Character & 0.43 & 0.38 & 2.86 & 0.66 \\
        Colt Express & Train & 0.11 & 0.08 & 1.48 & 0.07 \\
        Colt Express & Rounds & 0.14 & 0.11 & 1.77 & 0.19 \\
        \midrule
        Theme Park & All & 0.56 & 0.44 & 2.94 & 0.76\\
        Theme Park & Person Cards & 0.37 & 0.29 & 2.61 & 0.54\\
        Theme Park & Magic Cards & 0.43 & 0.31 & 2.73 & 0.71\\
        \midrule
        Catan & All & 0.70 & 0.51 & 3.13 & 0.81 \\
        Catan & Map & 0.50 & 0.31 & 2.55 & 0.49\\
        Catan & Dice & 0.14 & 0.11 & 1.86 & 0.28 \\
        \midrule
    \end{tabular}
    \caption{Results for each game and analysed source of randomness. `Constant' indicates what is held constant within each of the 100 sampled seeds; `All' corresponds to the data in Table~\ref{tab:Overall}. }
    \label{tab:Disentangling}
\end{table}

For Seven Wonders the distribution of the player boards has more impact on the game outcome than the shuffling of the three Age decks, although both are significant contributors to the total variation. 
Holding the board seed constant gives 55\% of tournaments outside the 99\% theoretical confidence bounds; 39\% are outliers when the card seed is fixed.

For Colt Express the impact on the game outcome is driven by the initial deal of each player's character (66\% outliers), with much smaller contributions from the Train (7\%) and Round (19\%)  decks.
The player boards in Seven Wonders and the players' characters in Colt Express are public information. This can lead to experienced players knowing before the game starts which players have a positional advantage, evidenced by player discussions for both games on forums~\cite{raimbault_colt_2016,clark_wonders_2021}.

The designer of Colt Express weighed in on this discussion, and clarified that perfect balance was not a key criterion, and was subsidiary to making each character uniquely interesting:
\begin{quote}
... balance issues were not my first focus. I wanted to get a fun game and I'm quite happy with the result. In Colt Express, I wanted everyone to have a fun time and not only the guy who will win.
I wanted to give each character a strong identity. Each ability is unique to make each player want to play at least one time each character of the game. Each ability needs you to play each character differently. So the conclusion is: I had to create quite unbalanced abilities.\cite{raimbault_colt_2016} 
\end{quote}

In Dominion the impact is dominated by the initial shuffle and the content of the first two hands. Fixing just this first shuffle gives 43\% outliers compared to only slightly more (56\%) across all variation. A game of Dominion using the recommended first game cards involves 5-10 reshuffles of each deck after the original one and these have less impact.
The overall impact of randomness in Dominion remains low in comparison to other imperfect information games looked at.

In Theme Park the total variation in outcome is more evenly balanced between the two components, with the Magic Card deck shuffle having a slightly larger impact; 71\% outliers and a trimmed span of 0.31 versus 54\% and 0.29 for Person cards.

The Person cards are all dealt at the start of the game and are public information. In this they are similar to the  boards in Seven Wonders and the characters in Colt Express.
Unlike those two games the Person cards do not have unique abilities and flavour. Each is just a list of 3 different rides the Person is keen to go on. The variation in outcome stems from synergies (or their lack) between the four cards that a player receives. The same ride appearing on 2 or 3 different cards, or rides across different cards being near to each other, makes it easier to visit the rides in the time available.

The Magic cards in contrast are all unique special abilities, and are only revealed during play for players to pick up and use. This means that any inbuilt advantage to one player is less visible at the start, and much of the fun of the game comes from the strong and varied effects of these cards.

As a result of this analysis, the publisher decided to reduce the variation arising from Person cards by having predefined sets of starting cards, each of which could be more balanced. They were much less concerned about the variation from Magic cards.

\subsection{Catan}\label{sect:ResultsCatan}
\begin{figure}
    \centering
    \includegraphics[scale=0.8]{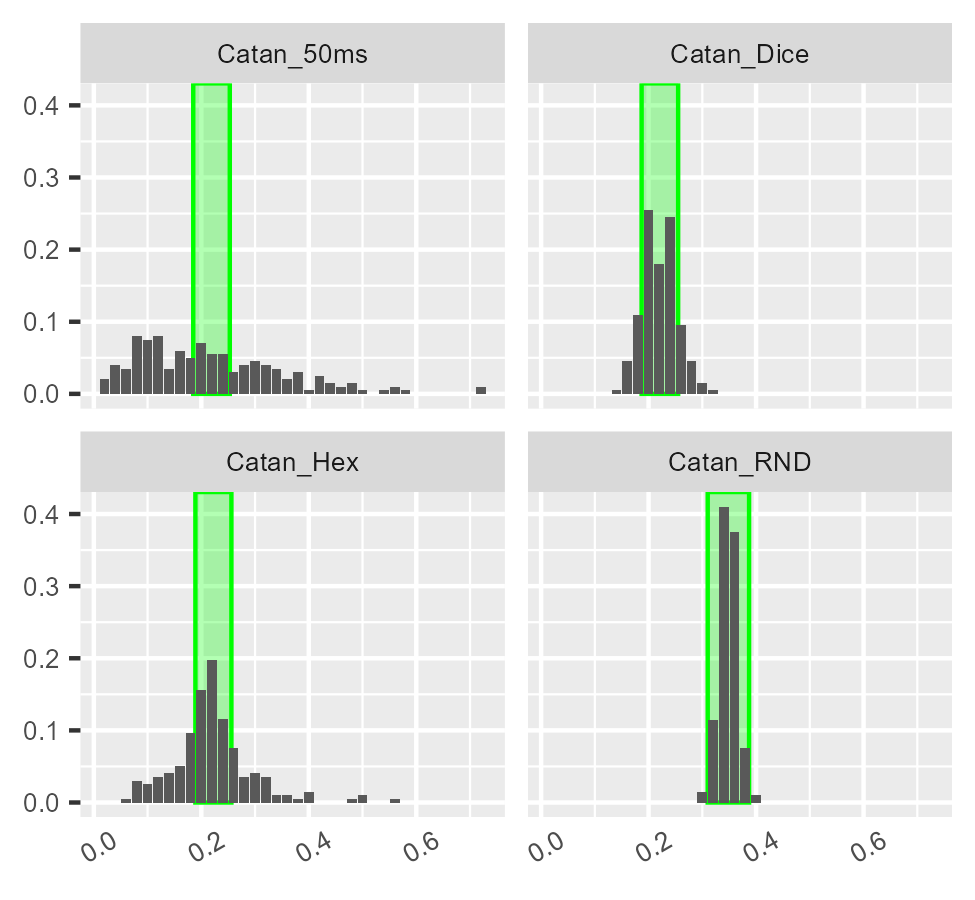}
    \caption{Results for Catan. With random players the stochasticity of the game has no impact. The other plots then show the relative impacts of the initial map set up and later dice rolls.}
    \label{fig:Catan}
\end{figure}

Catan is well known to be affected by the random set up of the board, but this was not evident in the original work~\cite{goodman_measuring_2024}. 
MCTS with a 50ms budget and no game-specific heuristics is very poor at the game, and this was believed to lie behind the lack of any variation in outcome across different random seeds.
To address this a game-specific state value function was trained from games between progressively better `expert' MCTS agents (with large budgets and earlier versions of the value function)~\cite{anthony_expert_2021}.

Using this improved agent gave much better results shown in Figure~\ref{fig:Catan} and Table~\ref{tab:Disentangling}. With random agents the seed has no impact on the outcome, as the agents are unable to exploit any benefits (as was the case in \cite{goodman_measuring_2024}).

There remains a caveat on the agent performance, with an unexpectedly strong first player disadvantage (22\% mean win rate for the first player). This is highly variable with the random seed, and 95\% of games have an underlying win rate between 4\% and 56\% (a trimmed span of 52\%).

This permitted analysis of the contributions from the initial board set up (fully visible at the start), and the sequence of dice throws during the game (unknowable in advance, but advantaging different hexes for resource generation).
Figure~\ref{fig:Catan} shows that the initial shuffle of hex board is much more important in affecting outcome than the sequence of dice rolls. 
The impact of the hex board layout is much reduced from the overall impact of randomness. This may be due to interaction effects between hex and dice seeds when both are fixed. 
In Catan the hex seed determines which hexes are valuable when a 6 (or any other number) is rolled, so this becomes more/less valuable based on how many 6s are actually rolled. This is controlled by the dice seed leading to synergies between them when both are fixed.

\section{Discussion}

Tables~\ref{tab:Overall} and~\ref{tab:Disentangling} show significant differences between games when measuring the mean win rate of the first player for a fixed random seed. 
This is not of itself `good' or `bad', and depends on the design goals. In the case of Colt Express the designer stated that the impact of the random allocation of characters is deliberate, 
or at least that balance is a low priority compared to flavour and varied player experience, 
A similar point has been made by designers of other board games~\cite{olotka_fair_2011,reiber_major_2021}. 

In Theme Park, these results helped prompt a re-design of one aspect of the game as having too strong an impact of relatively undifferentiated player set ups was not wanted, while the larger effect of the Magic card shuffle was acceptable.
These examples illustrate how the method proposed here can be used in practice to determine if game design goals are achieved and that the impact of randomness is within desired bounds.

Measuring the overall randomness of a game is distinct from identifying `strong' and `weak' cards or characters. One unbalanced character may strongly affect a small subset of games for example without affecting the holistic measure much. 
This holistic measure of randomness across the game also captures interactions between different elements, where the strength of a card depends on other cards in play. 

A high impact of randomness does not mean that a random player can do well in a strongly random game. Figure~\ref{fig:budgetVariance} disproves this and skill is often needed to exploit opportunities offered by chance.

All the various metrics tried show the same pattern across games. The Entropy measure seems least useful as it gives lower values for a game with peaks at $0\%$ or $100\%$ win rates, such as Poker, over games with narrower span but more central distribution. Both the Trimmed Span and Outlier count give a more robust single number.
Plans for future work to address weaknesses and follow-on questions are:
\begin{itemize}
    \item The agents used have deliberately been identical in terms of ability. We expect randomness to reduce the gap between asymmetric players, but this has not been tested. Do the same patterns occur for heterogeneous skill levels. If player 1 is \emph{better} than player 2, does that reduce the random effect?
    \item Only MCTS agents have been used.
Other techniques may be more suited to some of these games, and using a counterfactual-regret based approach could change the result of games like Poker~\cite{lanctot_search_2014}.
    \item Seed-picking. This work has looked at the distributional effect of random seeds. Curating individual seeds for games that have strongly skewed results can provide insight to the detailed causes of impact. Curating seeds with a balanced win rate can be helpful to better measure the relative performance of two agents, and a sequence of increasingly `hard' seeds could be used to build a curriculum for training reinforcement learning agents.
    \item Non-monotonic seeds have been identified for some games. Analysis of these is planned future work that may point to an interesting underlying feature of games.
    \item The first player win rate has been used to measure the impact of a random seed. This is an arbitrary choice and others may be of interest. For example in a three-player game a seed could have most of its impact on the relative success of players 2 and 3. Other game-specific measures of interest could whether a player ``Shot the Moon" in Hearts, or in Dominion which of the two different end-game conditions was met.
\end{itemize}

\section{Conclusion}
We have presented a technique to quantify the impact of the inherent randomness in a game on the outcome, and to disentangle the contributions to this from different sources of randomness.
The analysis of errors in experiments using agent win rate have shown that running mirror games for the same seed can usefully reduce practical experimental error, even if this is not formally quantifiable.

We have conducted a comparative analysis of 15 popular tabletop board and card games, and analysed the contribution of specific aspects of the game in selected cases. The insights these provide can be of use to the game designer as we show with a case study using the technique in a game currently in development by a commercial publisher.

Experiments with agents of varying skill supports the idea that for a given random seed increasing player skill can tend to a deterministic win for one player as any benefit is optimally exploited. This is especially true for games in which the results of the seed are visible to all players at game set up.

This technique is of general use as a tool for game design as well as for the analysis of existing games.
A game is not `bad' if randomness has a big impact on the game outcome (or vice versa). What matters is whether the level of impact is in line with the design objectives for the game.
The complaint that, ``the dice were against me", is sometimes a very valid one, and that is exactly as it should be.

\section*{Acknowledgment}
This work was funded by the EPSRC CDT in Intelligent Games and Game Intelligence (IGGI) EP/S022325/1. Gemini was used for some re-phrasing of text.
This research used Queen Mary's Apocrita HPC facility, supported by QMUL Research-IT. http://doi.org/10.5281/zenodo.438045
For the purpose of open access, the author(s) has applied a Creative Commons Attribution (CC BY) license to any Accepted Manuscript version arising.

\bibliographystyle{IEEEtran}
\bibliography{references}

\end{document}